\documentclass[letterpaper]{article} 
\usepackage{aaai2026}  
\usepackage{times}  
\usepackage{helvet}  
\usepackage{courier}  
\usepackage[hyphens]{url}  
\usepackage{graphicx} 
\usepackage{natbib}  
\usepackage{caption} 
\usepackage{algorithm}
\usepackage{algorithmic}
\newtheorem{theorem}{Theorem}
\usepackage{newfloat}
\usepackage{listings}
\DeclareCaptionStyle{ruled}{labelfont=normalfont,labelsep=colon,strut=off} 
\floatstyle{ruled}
\newfloat{listing}{tb}{lst}{}
\floatname{listing}{Listing}
\usepackage{amsmath}
\usepackage{amssymb}
\usepackage{array}
\usepackage{bm}
\usepackage{multirow}
\usepackage{booktabs}
\usepackage{siunitx}
\usepackage{rotating}
\title{KFS: KAN based adaptive Frequency Selection \\ learning architecture  for long term time series forecasting}
\author{
    Changning Wu\textsuperscript{\rm 1},
    Gao Wu\textsuperscript{\rm 1},
    Rongyao Cai\textsuperscript{\rm 1},
    Yong Liu\textsuperscript{*,1}
    ,Kexin Zhang\textsuperscript{*,1}
    ,
}
\affiliations{
    \textsuperscript{1}
Insititute of Cyber-Systems and Control, Zhejiang University, Hangzhou, China\\
\textsuperscript{*}Corresponding Author
    
}

\usepackage{bibentry}

\begin{document}

\maketitle
\begin{abstract}
Multi-scale decomposition architectures have emerged as predominant methodologies in time series forecasting. However, real-world time series exhibit noise interference across different scales, while heterogeneous information distribution among frequency components at varying scales leads to suboptimal multi-scale representation. Inspired by Kolmogorov-Arnold Networks (KAN) and Parseval's theorem, we propose a KAN based adaptive frequency Selection learning architecture (KFS) to address these challenges. This framework tackles prediction challenges stemming from cross-scale noise interference and complex pattern modeling through its FreK module, which performs energy-distribution-based dominant frequency selection in the spectral domain. Simultaneously, KAN enables sophisticated pattern representation while timestamp embedding alignment synchronizes temporal representations across scales. The feature mixing module then fuses scale-specific patterns with aligned temporal features. Extensive experiments across multiple real-world time series datasets demonstrate that KFS achieves state-of-the-art performance as a simple yet effective architecture.
\end{abstract}

\begin{links}
    \link{Code}{https://github.com/wcnExplosion/KFS-main}
\end{links}

\section{Introduction}
Time series forecasting (TSF) is applied in various significant domains, including finance ~\cite{huang2024generative}, traffic flow control ~\cite{JIANG2022117921}, and weather forecasting ~\cite{doi:10.1126/science.adi2336}. Recently, deep learning methods have driven continuous progress in the TSF field, with CNN-based ~\cite{donghao2024moderntcn}, MLP-based ~\cite{nie2023a}, and Transformer-based ~\cite{Zeng2022AreTE} approaches.

Due to real-world complexities, observed time series often exhibit intricate and diverse patterns. These interwoven patterns result in complex dependencies with substantial noise contamination, making it challenging to establish connections between historical data and future variations. To capture complex temporal patterns, increasing research focuses on leveraging prior knowledge to decompose time series into simpler components as the foundation for forecasting. For example, Autoformer~\cite{NEURIPS2021_bcc0d400}, DLinear~\cite{Zeng2022AreTE}, and FEDformer~\cite{zhou2022fedformer} decompose time series into trend and seasonal components. Building on this, TimeMixer~\cite{wang2024timemixer} further introduces multi-scale seasonal-trend decomposition, highlighting the importance of multi-scale data. Recent models like TimesNet~\cite{wu2023timesnet} and SparseTFT~\cite{lin2024sparsetsf} concentrate on decomposing long sequences into multiple shorter sub-sequences based on periodicity length. While these methods extract subsequences from diverse perspectives to capture critical information, the subsequences split directly from the original series inevitably retain substantial noise, leading to suboptimal problems. 

It is worth noting that time series contain multiple frequency components, including noise that interferes with model learning. This inherent noise affects different frequencies unevenly, causing lower signal-to-noise ratios at lower-amplitude frequencies and consequently impairing model predictive performance. Mitigating noise interference while blending diverse frequency components makes forecasting particularly challenging. The aforementioned decomposition methods inspire us to design a multi-scale frequency denoising hybrid framework capable of isolating different frequency components while filtering high signal-to-noise ratio data. However, heterogeneous frequency patterns introduce complex representational challenges, often yielding suboptimal results. Fortunately, Kolmogorov-Arnold Network (KAN)~\cite{liu2025kan} has recently gained significant attention in deep learning for its powerful data-fitting capability and flexibility, demonstrating potential to replace traditional MLPs. Compared to MLPs, KAN employs learnable activation functions that control its fitting capacity by adjusting basis functions. Moreover, TimeKAN~\cite{huang2025timekan}, a KAN based method, has achieved SOTA performance in multiple datasets, demonstrating the remarkable potential of KAN for temporal feature representation. These considerations motivate us to explore KAN for representing patterns across different frequencies, thereby providing more information for forecasting.

Inspired by these observations, we propose KAN based adaptive frequency Selection learning architecture (KFS) to address forecasting challenges arising from noise and mixed data pattern. Specifically, KFS first decomposes components within the data via moving averages. Subsequently, the FreK module performs frequency selection at multiple scales to denoise the data, utilizing KAN to learn scale-specific temporal features from the denoised data. Finally, the hybrid module aligns and fuses timestamp embeddings from the look back window with corresponding scale representations, achieving temporal representation alignment and integration across scales to precisely model temporal features. Features from different scales are aggregated via averaging and projected to the desired forecast horizon through simple linear mapping. With our meticulously designed architecture, KFS achieves state-of-the-art performance in long-term time series forecasting tasks across multiple real-world datasets.

Our contributions can be summarized as follows:

\begin{itemize}
    \item We designed an energy-distribution-based frequency selection method that effectively extracts components with higher signal-to-noise ratios. The resulting FreK module reduces noise impact and enables efficient modeling.
    \item We introduced a simple yet effective forecasting model KFS, and developed a Mixing Block that aligns and fuses multi-scale time series with corresponding timestamps.
    \item Comprehensive experiments demonstrate that our KFS achieves state-of-the-art performance in long-term forecasting tasks across multiple datasets while exhibiting exceptional efficiency.
\end{itemize}

\section{Related Works}
\subsection{Time Series Forecasting}
In recent years, deep learning approaches for TSF have gained significant attention, mainly including CNN-based, MLP-based, and Transformer-based methodologies.

CNN-based methods focus on extracting temporal feature representations through convolutional operations. For example, MICN ~\cite{wang2023micn} and TimesNet ~\cite{wu2023timesnet} enhance the accuracy of sequence modeling by strategically adjusting the receptive fields of their architectures. Transformer-based approaches, while contrasting with CNN methods, exhibit substantially larger receptive fields. PatchTST ~\cite{nie2023a} improves the capture of local patterns by segmenting input data into patches, while Crossformer ~\cite{zhang2023crossformer} specializes in mining cross-variable dependencies. However, Transformer-based models face challenges stemming from computational complexity due to their massive parameterization. In this situation, MLP-based methods secure their position in TSF through lightweight architectures. FITS ~\cite{xu2024fits} introduces novel linear projections to reduce input complexity, requiring merely 10K parameters. However, constrained by their parameterization, MLP-based approaches struggle to effectively extract and fuse diverse data modalities.

Unlike the aforementioned methods, this paper enhance data quality through spectral filtering strategies and integrate a multi-scale framework to extract temporal  representations, achieving significantly improved accuracy in long term Time Series Forecasting.

\subsection{Multi-Scale Architecture for TSF}

In the field of TSF, extensive research has explored multi-scale architectures. TimeMixer ~\cite{wang2024timemixer} pioneered their application in TSF through decomposing multi-scale time series. MICN ~\cite{wang2023micn} extended multi-scale processing to convolutional layers, enabling efficient representation of seasonal patterns. Building on these advances, this work leverages a multi-scale framework to capture hierarchical information, proposing KFS's novel multi-pathway integration framework. By distinctly capturing temporal representations and physical timestamp embeddings, then fusing these components, KFS achieves enhanced precision in time series forecasting.

\subsection{Kolmogorov-Arnold Network}

The Kolmogorov-Arnold representation theorem establishes that any multivariate continuous function can be expressed as a composition of univariate functions and additive operations. Using this theorem, KAN ~\cite{liu2025kan} introduces a novel network architecture that supplants traditional MLPs. Unlike MLPs with fixed activation functions, KAN incorporates learnable activation functions. This flexibility positions KAN as a promising alternative to MLPs.

Initial implementations of KAN faced computational bottlenecks due to the excessive complexity of B-spline sampling, hindering broader adoption. To address this limitation, subsequent research explored alternative basis functions, rKAN ~\cite{aghaei2024rkanrationalkolmogorovarnoldnetworks} investigates rational functions as basis functions, FastKAN ~\cite{li2024kolmogorovarnoldnetworksradialbasis} accelerates computation using Gaussian radial basis functions to approximate third-order B-spline functions.

Furthermore, KAN has been adopted across diverse domains as a substitute for MLP. Convolutional KAN~\cite{bodner2025convolutionalkolmogorovarnoldnetworks} replaces conventional kernels with learnable spline functions. KAT~\cite{yang2025kolmogorovarnold} integrates KAN layers into Transformer architectures, demonstrating impressive accuracy in multiple computer vision tasks. This paper proposes to introduce KAN to TSF and explore its potential in representing temporal data patterns.

\section{Preliminary}
\subsection{Motivation}
In the physical world, time series data originate from sensors on physical devices or recordings of real-world relationships. These measurements inherently contain varying levels of noise interference due to factors including acquisition methods, mechanical transmission processes, and recording mechanisms. This noise significantly compromises the results of time series analysis tasks, particularly forecasting and anomaly detection. Consequently, developing methodologies to mitigate noise-induced distortions becomes imperative to enhance the representation of temporal patterns. This paper addresses this challenge through the view of multivariate time series forecasting.

We formally decompose the forecasting problem into two fundamental questions.

\begin{enumerate}
    \item \textbf{How can we effectively reduce noise impact on both data and predictive models?}  
    \item \textbf{How can we explicitly extract intrinsic information from given time series?}  
\end{enumerate}
\begin{figure*}[t]
\centering
\includegraphics[width=1\textwidth]{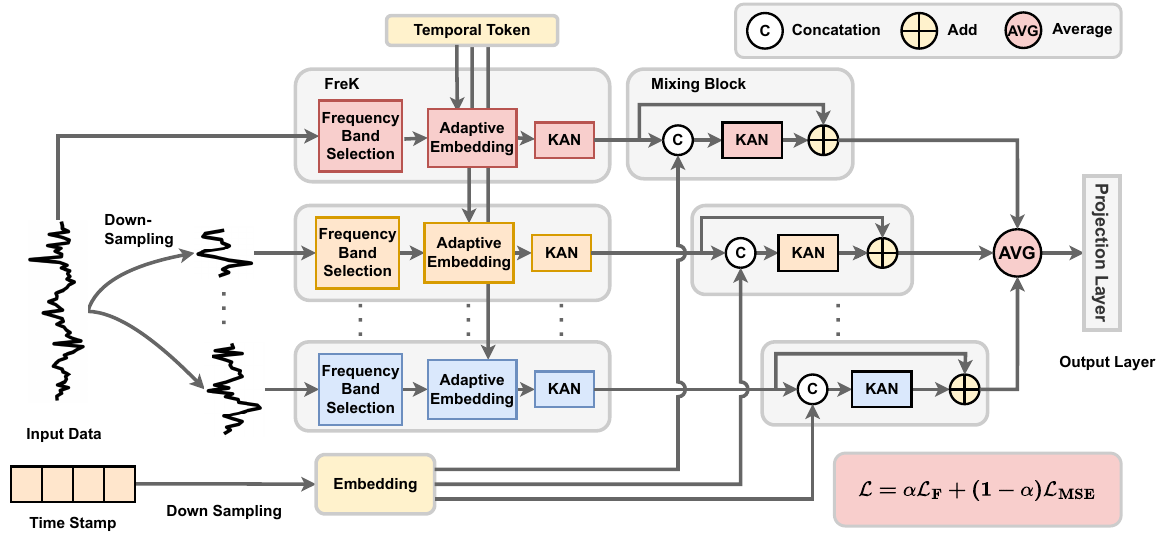} 
\caption{Overall structure of the proposed KFS. Multi-scale architecture decomposes time series. The KANs are seamlessly integrated within the model framework. FreK select the dominant frequency based on energy distribution and represent the temporal pattern. Mixing Block align temporal representation with its time stamp.}
\label{fig:KFS}
\end{figure*}
For the first question, we begin by assuming that the data primarily contains channel-wise independent additive white Gaussian noise. The primary mitigation approach for such noise traditionally requires prior knowledge of the noise distribution, which introduces additional domain-specific assumptions and hinders real-world applicability. However, the spectral uniformity of Gaussian white noise in the frequency domain motivates our solution: By selecting frequency bands with concentrated energy as the dominant temporal features, we reconstruct the time series within a bounded error margin, effectively attenuating noise. This principle is formalized in the following theorems.
\begin{theorem}[Parseval's Theorem]
For a discrete signal \( y \in \mathbb{R}^L \) and its DFT \( Y \in \mathbb{C}^{L/2+1} \), the energy satisfies:
\begin{equation}
\sum_{t=0}^{L-1} |y(t)|^2 = \frac{1}{L} \sum_{k=0}^{L/2} |Y[k]|^2.
\end{equation}
\label{Parseval}
\end{theorem}
Theorem 1 states that the total energy of a time series is equivalent in the frequency domain and the time domain. Therefore, by processing the time series in the frequency domain and converting it back to the time domain, the information of the original time series can be preserved. This foundation allows us to formalize \textbf{Theorem 2}, with its complete proof detailed in Appendix.
\begin{theorem}
Let observed time series \( y = y_0 + n \), where \( n \sim \mathcal{N}(0, \sigma^2 I) \), and $y_0$ donates original times series. After DFT, there exist \( K \in \mathbb{N}^+ \) and \( \epsilon > 0 \) such that the sparse reconstruction \( \tilde{y} \) from the top-$K$ frequencies of \( Y \) satisfies:
\begin{equation}
\|\tilde{y} - y_0\|_2 < \epsilon.
\end{equation}
\label{Distribution}
\end{theorem}
Theorem 2 proposes that by filtering the dominant frequency bands of a time series, the proportion of noise can be reduced, thereby enhancing the quality of time series.

For the second question, we draw upon existing research to carefully design a KAN-based network architecture under the channel independence assumption, integrated with a multi-scale time series mixing framework.

\subsection{Multiscale Time Series Processing}

In long time series forecasting, temporal sequences can capture information from multiple scales by down sampling, thereby enhancing prediction accuracy. For an input time series $X \in \mathbb{R}^{L \times C}$, we generate multi-scale sequences through down sampling. Specifically, for each coarser-grained subsequence $X_{i+1}$, it is derived from the finer-grained subsequence $X_i$ at the preceding level by applying average pooling. We then sequentially obtain a collection of time series $\mathbb{X} = \{ x_0, x_1, ..., x_m \}$ across m scales, where each $x_i \in \mathbb{R} ^ {\frac{L}{D^i} / \times C}$ and $D$ donates the window size of average pooling. The down sampling process used in our work is shown as below:
\begin{equation}
x_{i+1} = \text{AvgPool}(x_i)
\end{equation}

This technique has been extensively adopted in time series forecasting models and has demonstrated improved predictive accuracy along with enhanced modeling capabilities.

\section{Method}

\subsection{Overview of the architecture}

The core challenge lies in resolving sequence modeling for channel-independent information while effectively reducing the influence of noise. To address this, we propose a simple yet effective architecture, the \textbf{K}AN based adaptive \textbf{F}requency \textbf{S}election learning architecture  (KFS), which improves prediction accuracy by organically integrating KAN to capture multiscale channel-independent features and temporal representation. The overall architecture of KFS is depicted in Figure ~\ref{fig:KFS}. Specifically, it consists of two key components: a \textbf{Fre}quency \textbf{K}-top Selection (\textbf{FreK}) Module and a Mixing Block. Detailed descriptions of each module are deferred to the following sections.

\begin{table*}[t]
    \centering
    \small
\setlength{\tabcolsep}{1.4pt}
\begin{tabular}{*{22}{c}} 
\toprule
\multicolumn{2}{c}{Models}
 & \multicolumn{2}{c}{KFS(Ours)} & \multicolumn{2}{c}{TimeXer} & \multicolumn{2}{c}{TimeMixer} & \multicolumn{2}{c}{iTransformer} & 
\multicolumn{2}{c}{PatchTST} & \multicolumn{2}{c}{TimesNet} & \multicolumn{2}{c}{MICN} & \multicolumn{2}{c}{DLinear} & 
\multicolumn{2}{c}{FiLM} &\multicolumn{2}{c}{Time-FFM} \\
\cmidrule(lr){3-4} \cmidrule(lr){5-6} \cmidrule(lr){7-8} \cmidrule(lr){9-10} 
\cmidrule(lr){11-12} \cmidrule(lr){13-14} \cmidrule(lr){15-16} \cmidrule(lr){17-18} \cmidrule(lr){19-20} \cmidrule(lr){21-22}

\multicolumn{2}{c}{Metric} &MSE &\multicolumn{1}{c|}{MAE} &MSE &\multicolumn{1}{c|}{MAE} &MSE &\multicolumn{1}{c|}{MAE} &MSE &\multicolumn{1}{c|}{MAE} &MSE &\multicolumn{1}{c|}{MAE} &MSE &\multicolumn{1}{c|}{MAE} &MSE &\multicolumn{1}{c|}{MAE} &MSE &\multicolumn{1}{c|}{MAE} &MSE &\multicolumn{1}{c|}{MAE}  &MSE &\multicolumn{1}{c}{MAE} \\
\midrule
\multirow{5}{*}{\begin{turn}{90}Weather\end{turn}} 
&\multicolumn{1}{|c}{96} 
&\multicolumn{1}{|c}{ {\underline{0.159}} } &  {\textbf{0.205}} 
&\multicolumn{1}{|c}{ {\textbf{0.157}} }  &  {\textbf{0.205} }
&\multicolumn{1}{|c}{0.163} & 0.209 &\multicolumn{1}{|c}{0.174} & 0.214
&\multicolumn{1}{|c}{0.186} & 0.227
&\multicolumn{1}{|c}{0.172} & 0.220
&\multicolumn{1}{|c}{0.198} & 0.261
&\multicolumn{1}{|c}{0.195} & 0.252
&\multicolumn{1}{|c}{0.195} & 0.236
&\multicolumn{1}{|c}{0.191} & 0.230
\\
&\multicolumn{1}{|c}{192} 
&\multicolumn{1}{|c}{ {\underline{0.207}}} &  {\underline{0.249}}
&\multicolumn{1}{|c}{ {\textbf{0.204}}} &  {\textbf{0.247} }
&\multicolumn{1}{|c}{0.211} & 0.254 
&\multicolumn{1}{|c}{0.221} & 0.254
&\multicolumn{1}{|c}{0.234} & 0.265
&\multicolumn{1}{|c}{0.219} & 0.261
&\multicolumn{1}{|c}{0.239} & 0.299
&\multicolumn{1}{|c}{0.237} & 0.295
&\multicolumn{1}{|c}{0.239} & 0.271
&\multicolumn{1}{|c}{0.236} & 0.267
\\
&\multicolumn{1}{|c}{336} 
&\multicolumn{1}{|c}{ {\underline{0.262}}} &  {\textbf{0.288}}
&\multicolumn{1}{|c}{ {\textbf{0.261}}} &  {\underline{0.290}}
&\multicolumn{1}{|c}{0.263} &0.293
&\multicolumn{1}{|c}{0.278} & 0.296
&\multicolumn{1}{|c}{0.284} & 0.301
&\multicolumn{1}{|c}{0.280} & 0.306
&\multicolumn{1}{|c}{0.285} & 0.336
&\multicolumn{1}{|c}{0.282} & 0.331
&\multicolumn{1}{|c}{0.289} & 0.306
&\multicolumn{1}{|c}{0.289} & 0.303
\\
&\multicolumn{1}{|c}{720} 
&\multicolumn{1}{|c}{0.345} &  {\underline{0.342}}
&\multicolumn{1}{|c}{ {\textbf{0.340}}} &  {\textbf{0.341} }
&\multicolumn{1}{|c}{ {\underline{0.344}}} & 0.348 
&\multicolumn{1}{|c}{0.358} & 0.347
&\multicolumn{1}{|c}{0.356} & 0.349
&\multicolumn{1}{|c}{0.365} & 0.359
&\multicolumn{1}{|c}{0.351} & 0.388
&\multicolumn{1}{|c}{0.345} & 0.382
&\multicolumn{1}{|c}{0.360} & 0.351
&\multicolumn{1}{|c}{0.362} & 0.350
\\

\cmidrule(r){2-22}
&\multicolumn{1}{|c}{Avg} 
&\multicolumn{1}{|c}{ {\underline{0.243}}} &  {\textbf{0.271}}
&\multicolumn{1}{|c}{ {\textbf{0.241}}} &  {\textbf{0.271}}
&\multicolumn{1}{|c}{0.245} &  {\underline{0.276}}
&\multicolumn{1}{|c}{0.256} & 0.278
&\multicolumn{1}{|c}{0.265} & 0.285
&\multicolumn{1}{|c}{0.259} & 0.287
&\multicolumn{1}{|c}{0.268} & 0.321
&\multicolumn{1}{|c}{0.265} & 0.315
&\multicolumn{1}{|c}{0.271} & 0.290
&\multicolumn{1}{|c}{0.270} & 0.288

\\
\midrule
\multirow{5}{*}{\begin{turn}{90}ETTh1\end{turn}} 
&\multicolumn{1}{|c}{96} 
&\multicolumn{1}{|c}{ {\textbf{0.368}}} &  {\textbf{0.397}}
&\multicolumn{1}{|c}{ {\underline{0.382}}} & 0.403 
&\multicolumn{1}{|c}{0.385} & 0.402
&\multicolumn{1}{|c}{0.386} & 0.405
&\multicolumn{1}{|c}{0.460} & 0.447
&\multicolumn{1}{|c}{0.384} & 0.402
&\multicolumn{1}{|c}{0.426} & 0.446
&\multicolumn{1}{|c}{0.395} & 0.407
&\multicolumn{1}{|c}{0.438} & 0.433
&\multicolumn{1}{|c}{0.385} &  {\underline{0.400}}
\\
&\multicolumn{1}{|c}{192} 
&\multicolumn{1}{|c}{ {\textbf{0.425}}} &  {\textbf{0.426}}
&\multicolumn{1}{|c}{ {\underline{0.429}}} & 0.435
&\multicolumn{1}{|c}{0.443} &  0.430
&\multicolumn{1}{|c}{0.441} & 0.436
&\multicolumn{1}{|c}{0.512} & 0.477
&\multicolumn{1}{|c}{0.436} &  {\underline{0.429}}
&\multicolumn{1}{|c}{0.454} & 0.464
&\multicolumn{1}{|c}{0.446} & 0.441
&\multicolumn{1}{|c}{0.494} & 0.466
&\multicolumn{1}{|c}{0.439} & 0.430
\\
&\multicolumn{1}{|c}{336} 
&\multicolumn{1}{|c}{ {\textbf{0.467}}} &  {\textbf{0.446}}
&\multicolumn{1}{|c}{ {\underline{0.468}}} &  {\underline{0.448}}
&\multicolumn{1}{|c}{0.512} & 0.470
&\multicolumn{1}{|c}{0.487} & 0.458
&\multicolumn{1}{|c}{0.546} & 0.496
&\multicolumn{1}{|c}{0.491} & 0.469
&\multicolumn{1}{|c}{0.493} & 0.487
&\multicolumn{1}{|c}{0.489} & 0.467
&\multicolumn{1}{|c}{0.547} & 0.495
&\multicolumn{1}{|c}{0.480} & 0.449
\\
&\multicolumn{1}{|c}{720} 
&\multicolumn{1}{|c}{ {\textbf{0.454}}} &  {\underline{0.458}}
&\multicolumn{1}{|c}{0.469} &  0.461
&\multicolumn{1}{|c}{0.497} & 0.476
&\multicolumn{1}{|c}{0.503} & 0.491
&\multicolumn{1}{|c}{0.544} & 0.517
&\multicolumn{1}{|c}{0.521} & 0.500
&\multicolumn{1}{|c}{0.526} & 0.526
&\multicolumn{1}{|c}{0.513} & 0.510
&\multicolumn{1}{|c}{0.586} & 0.538
&\multicolumn{1}{|c}{ {\underline{0.462}}} &  {\textbf{0.456}}
\\

\cmidrule(r){2-22}
&\multicolumn{1}{|c}{Avg} 
&\multicolumn{1}{|c}{ {\textbf{0.428}}} &  {\textbf{0.431}}
&\multicolumn{1}{|c}{ {\underline{0.437}}} & 0.437
&\multicolumn{1}{|c}{0.459} & 0.444
&\multicolumn{1}{|c}{0.454} & 0.447
&\multicolumn{1}{|c}{0.516} & 0.484
&\multicolumn{1}{|c}{0.458} & 0.450
&\multicolumn{1}{|c}{0.475} & 0.480
&\multicolumn{1}{|c}{0.461} & 0.457
&\multicolumn{1}{|c}{0.516} & 0.483
&\multicolumn{1}{|c}{0.442} &  {\underline{0.434}}

\\
\midrule
\multirow{5}{*}{\begin{turn}{90}ETTh2\end{turn}} 
&\multicolumn{1}{|c}{96} 
&\multicolumn{1}{|c}{ {\textbf{0.280}}} &  {\textbf{0.334}}
&\multicolumn{1}{|c}{ {\underline{0.286}}} &  {\underline{0.338}}
&\multicolumn{1}{|c}{0.289} & 0.342
&\multicolumn{1}{|c}{0.297} & 0.349
&\multicolumn{1}{|c}{0.308} & 0.355
&\multicolumn{1}{|c}{0.340} & 0.374
&\multicolumn{1}{|c}{0.372} & 0.424
&\multicolumn{1}{|c}{0.340} & 0.394
&\multicolumn{1}{|c}{0.322} & 0.364
&\multicolumn{1}{|c}{0.301} & 0.351
\\
&\multicolumn{1}{|c}{192} 
&\multicolumn{1}{|c}{ {\textbf{0.362}}} &  {\textbf{0.387}}
&\multicolumn{1}{|c}{ {\underline{0.363}}} &  {\underline{0.389}}
&\multicolumn{1}{|c}{0.378} &  0.397
&\multicolumn{1}{|c}{0.380} & 0.400
&\multicolumn{1}{|c}{0.393} & 0.405
&\multicolumn{1}{|c}{0.402} & 0.414
&\multicolumn{1}{|c}{0.492} & 0.492
&\multicolumn{1}{|c}{0.482} & 0.479
&\multicolumn{1}{|c}{0.405} & 0.414
&\multicolumn{1}{|c}{0.378} & 0.397
\\
&\multicolumn{1}{|c}{336} 
&\multicolumn{1}{|c}{ {\textbf{0.406}}} &  {\textbf{0.421}}
&\multicolumn{1}{|c}{ {\underline{0.414}}} &  {\underline{0.423}}
&\multicolumn{1}{|c}{0.432} & 0.434
&\multicolumn{1}{|c}{0.428} & 0.432
&\multicolumn{1}{|c}{0.427} & 0.436
&\multicolumn{1}{|c}{0.452} & 0.452
&\multicolumn{1}{|c}{0.607} & 0.555
&\multicolumn{1}{|c}{0.591} & 0.541
&\multicolumn{1}{|c}{0.435} & 0.445
&\multicolumn{1}{|c}{0.422} & 0.431
\\
&\multicolumn{1}{|c}{720} 
&\multicolumn{1}{|c}{ {\underline{0.423}}} &  {\underline{0.435}}
&\multicolumn{1}{|c}{ {\textbf{0.408}}} &   {\textbf{0.432}}
&\multicolumn{1}{|c}{0.464} & 0.464
&\multicolumn{1}{|c}{0.427} & 0.445
&\multicolumn{1}{|c}{0.436} & 0.450
&\multicolumn{1}{|c}{0.462} & 0.468
&\multicolumn{1}{|c}{0.824} & 0.655
&\multicolumn{1}{|c}{0.839} & 0.661
&\multicolumn{1}{|c}{0.445} & 0.457
&\multicolumn{1}{|c}{0.427} & 0.444
\\

\cmidrule(r){2-22}
&\multicolumn{1}{|c}{Avg} 
&\multicolumn{1}{|c}{ {\textbf{0.367}}} &  {\textbf{0.394}}
&\multicolumn{1}{|c}{ {\textbf{0.367}}} &  {\underline{0.396}}
&\multicolumn{1}{|c}{0.390} & 0.409
&\multicolumn{1}{|c}{0.383} & 0.407
&\multicolumn{1}{|c}{0.391} & 0.441
&\multicolumn{1}{|c}{0.414} & 0.427
&\multicolumn{1}{|c}{0.574} & 0.531
&\multicolumn{1}{|c}{0.563} & 0.519
&\multicolumn{1}{|c}{0.402} & 0.420
&\multicolumn{1}{|c}{ {\underline{0.382}}} & 0.406

\\

\midrule
\multirow{5}{*}{\begin{turn}{90}ETTm1\end{turn}} 
&\multicolumn{1}{|c}{96} 
&\multicolumn{1}{|c}{ {\textbf{0.314}}} &  {\textbf{0.354}}
&\multicolumn{1}{|c}{0.318} &  {\underline{0.356}}
&\multicolumn{1}{|c}{ {\underline{0.317}}} &  {\underline{0.356}}
&\multicolumn{1}{|c}{0.334} & 0.368
&\multicolumn{1}{|c}{0.352} & 0.374
&\multicolumn{1}{|c}{0.338} & 0.375
&\multicolumn{1}{|c}{0.365} & 0.387
&\multicolumn{1}{|c}{0.346} & 0.374
&\multicolumn{1}{|c}{0.353} & 0.370
&\multicolumn{1}{|c}{0.336} & 0.369
\\
&\multicolumn{1}{|c}{192} 
&\multicolumn{1}{|c}{ {\textbf{0.358}}} &  {\textbf{0.378}}
&\multicolumn{1}{|c}{ {\underline{0.362}}} &  {\underline{0.383}}
&\multicolumn{1}{|c}{0.367} & 0.384
&\multicolumn{1}{|c}{0.377} & 0.391
&\multicolumn{1}{|c}{0.390} & 0.393
&\multicolumn{1}{|c}{0.374} & 0.387
&\multicolumn{1}{|c}{0.403} & 0.408
&\multicolumn{1}{|c}{0.382} & 0.391
&\multicolumn{1}{|c}{0.389} & 0.387
&\multicolumn{1}{|c}{0.378} & 0.389
\\
&\multicolumn{1}{|c}{336} 
&\multicolumn{1}{|c}{ {\textbf{0.388}}} &  {\textbf{0.398}}
&\multicolumn{1}{|c}{0.395} & 0.407
&\multicolumn{1}{|c}{ {\underline{0.391}}} &  {\underline{0.406}}
&\multicolumn{1}{|c}{0.426} & 0.420
&\multicolumn{1}{|c}{0.421} & 0.414
&\multicolumn{1}{|c}{0.410} & 0.411
&\multicolumn{1}{|c}{0.436} & 0.431
&\multicolumn{1}{|c}{0.415} & 0.415
&\multicolumn{1}{|c}{0.421} & 0.408
&\multicolumn{1}{|c}{0.411} & 0.410
\\
&\multicolumn{1}{|c}{720} 
&\multicolumn{1}{|c}{0.460} &  {\underline{0.446}}
&\multicolumn{1}{|c}{ {\textbf{0.452}}} &  {\textbf{0.441}}
&\multicolumn{1}{|c}{ {\underline{0.454}}} &  {\textbf{0.441}}
&\multicolumn{1}{|c}{0.491} & 0.459
&\multicolumn{1}{|c}{0.462} & 0.449
&\multicolumn{1}{|c}{0.478} & 0.450
&\multicolumn{1}{|c}{0.489} & 0.462
&\multicolumn{1}{|c}{0.473} & 0.451
&\multicolumn{1}{|c}{0.481} &  {\textbf{0.441}}
&\multicolumn{1}{|c}{0.469} &  {\textbf{0.441}}
\\

\cmidrule(r){2-22}
&\multicolumn{1}{|c}{Avg} 
&\multicolumn{1}{|c}{ {\textbf{0.380}}} &  {\textbf{0.394}}
&\multicolumn{1}{|c}{ {\underline{0.382}}} &  {\underline{0.397}}
&\multicolumn{1}{|c}{ {\underline{0.382}}} &  {\underline{0.397}}
&\multicolumn{1}{|c}{0.407} & 0.410
&\multicolumn{1}{|c}{0.406} & 0.407
&\multicolumn{1}{|c}{0.400} & 0.406
&\multicolumn{1}{|c}{0.423} & 0.422
&\multicolumn{1}{|c}{0.404} & 0.408
&\multicolumn{1}{|c}{0.412} & 0.402
&\multicolumn{1}{|c}{0.399} & 0.402

\\
\midrule
\multirow{5}{*}{\begin{turn}{90}ETTm2\end{turn}} 
&\multicolumn{1}{|c}{96} 
&\multicolumn{1}{|c}{ {\underline{0.173}}} &  {\textbf{0.253}}
&\multicolumn{1}{|c}{ {\textbf{0.171}}} &  {\underline{0.256}}
&\multicolumn{1}{|c}{0.175} & 0.257
&\multicolumn{1}{|c}{0.180} & 0.264
&\multicolumn{1}{|c}{0.183} & 0.270
&\multicolumn{1}{|c}{0.187} & 0.267
&\multicolumn{1}{|c}{0.197} & 0.296
&\multicolumn{1}{|c}{0.193} & 0.293
&\multicolumn{1}{|c}{0.183} & 0.266
&\multicolumn{1}{|c}{0.181} & 0.267
\\
&\multicolumn{1}{|c}{192} 
&\multicolumn{1}{|c}{ {\textbf{0.236}}} &  {\textbf{0.295}}
&\multicolumn{1}{|c}{ {\underline{0.237}}} &  {\underline{0.299}}
&\multicolumn{1}{|c}{0.240} &  0.302
&\multicolumn{1}{|c}{0.250} & 0.309
&\multicolumn{1}{|c}{0.255} & 0.314
&\multicolumn{1}{|c}{0.249} & 0.309
&\multicolumn{1}{|c}{0.284} & 0.361
&\multicolumn{1}{|c}{0.284} & 0.361
&\multicolumn{1}{|c}{0.248} & 0.305
&\multicolumn{1}{|c}{0.247} & 0.308
\\
&\multicolumn{1}{|c}{336} 
&\multicolumn{1}{|c}{ {\textbf{0.291}}} &  {\textbf{0.332}}
&\multicolumn{1}{|c}{ {\underline{0.296}}} &  {\underline{0.338}}
&\multicolumn{1}{|c}{0.303} & 0.343
&\multicolumn{1}{|c}{0.311} & 0.348
&\multicolumn{1}{|c}{0.309} & 0.347
&\multicolumn{1}{|c}{0.321} & 0.351
&\multicolumn{1}{|c}{0.381} & 0.429
&\multicolumn{1}{|c}{0.382} & 0.429
&\multicolumn{1}{|c}{0.309} & 0.343
&\multicolumn{1}{|c}{0.309} & 0.347
\\
&\multicolumn{1}{|c}{720} 
&\multicolumn{1}{|c}{ {\underline{0.395}}} &  {\underline{0.395}}
&\multicolumn{1}{|c}{ {\textbf{0.392}}} &   {\textbf{0.394}}
&\multicolumn{1}{|c}{ {\textbf{0.392}}} & 0.396
&\multicolumn{1}{|c}{0.412} & 0.407
&\multicolumn{1}{|c}{0.412} & 0.404
&\multicolumn{1}{|c}{0.408} & 0.403
&\multicolumn{1}{|c}{0.549} & 0.522
&\multicolumn{1}{|c}{0.558} & 0.525
&\multicolumn{1}{|c}{0.410} & 0.400
&\multicolumn{1}{|c}{0.406} & 0.404
\\

\cmidrule(r){2-22}
&\multicolumn{1}{|c}{Avg} 
&\multicolumn{1}{|c}{ {\textbf{0.274}}} & {\textbf{0.319}}
&\multicolumn{1}{|c}{ {\textbf{0.274}}} &  {\underline{0.322}}
&\multicolumn{1}{|c}{ {\underline{0.277}}} & 0.324
&\multicolumn{1}{|c}{0.288} & 0.332
&\multicolumn{1}{|c}{0.290} & 0.334
&\multicolumn{1}{|c}{0.291} & 0.333
&\multicolumn{1}{|c}{0.353} & 0.402
&\multicolumn{1}{|c}{0.354} & 0.402
&\multicolumn{1}{|c}{0.288} & 0.328
&\multicolumn{1}{|c}{0.286} & 0.332

\\
\midrule
\multirow{5}{*}{\begin{turn}{90}Electricity\end{turn}} 
&\multicolumn{1}{|c}{96} 
&\multicolumn{1}{|c}{ {\underline{0.148}}} &  {\textbf{0.238}}
&\multicolumn{1}{|c}{ {\textbf{0.140}}} &  {\underline{0.242}}
&\multicolumn{1}{|c}{0.153} & 0.245
&\multicolumn{1}{|c}{ {\underline{0.148}}} &  {\underline{0.240}}
&\multicolumn{1}{|c}{0.190} & 0.296
&\multicolumn{1}{|c}{0.168} & 0.272
&\multicolumn{1}{|c}{0.180} & 0.293
&\multicolumn{1}{|c}{0.210} & 0.302
&\multicolumn{1}{|c}{0.198} & 0.274
&\multicolumn{1}{|c}{0.198} & 0.282
\\
&\multicolumn{1}{|c}{192} 
&\multicolumn{1}{|c}{0.164} &  {\textbf{0.253}}
&\multicolumn{1}{|c}{ {\textbf{0.157}}} &  {\underline{0.256}}
&\multicolumn{1}{|c}{0.166} &  0.257
&\multicolumn{1}{|c}{ {\underline{0.162}}} & 0.253
&\multicolumn{1}{|c}{0.199} & 0.304
&\multicolumn{1}{|c}{0.184} & 0.289
&\multicolumn{1}{|c}{0.189} & 0.302
&\multicolumn{1}{|c}{0.210} & 0.305
&\multicolumn{1}{|c}{0.198} & 0.278
&\multicolumn{1}{|c}{0.199} & 0.285
\\
&\multicolumn{1}{|c}{336} 
&\multicolumn{1}{|c}{0.181} &  {\textbf{0.274}}
&\multicolumn{1}{|c}{ {\textbf{0.176}}} & 0.275
&\multicolumn{1}{|c}{0.185} & 0.275
&\multicolumn{1}{|c}{ {\underline{0.178}}} & 0.269
&\multicolumn{1}{|c}{0.217} & 0.319
&\multicolumn{1}{|c}{0.198} & 0.300
&\multicolumn{1}{|c}{0.198} & 0.312
&\multicolumn{1}{|c}{0.223} & 0.319
&\multicolumn{1}{|c}{0.217} & 0.300
&\multicolumn{1}{|c}{0.212} & 0.298
\\
&\multicolumn{1}{|c}{720} 
&\multicolumn{1}{|c}{0.219} &  {\textbf{0.306}}
&\multicolumn{1}{|c}{ {\textbf{0.211}}} &   {\textbf{0.306}}
&\multicolumn{1}{|c}{0.224} &  {\underline{0.312}}
&\multicolumn{1}{|c}{0.225} & 0.317
&\multicolumn{1}{|c}{0.258} & 0.352
&\multicolumn{1}{|c}{0.220} & 0.320
&\multicolumn{1}{|c}{ {\underline{0.217}}} & 0.330
&\multicolumn{1}{|c}{0.258} & 0.350
&\multicolumn{1}{|c}{0.278} & 0.356
&\multicolumn{1}{|c}{0.253} & 0.330
\\

\cmidrule(r){2-22}
&\multicolumn{1}{|c}{Avg} 
&\multicolumn{1}{|c}{ {\underline{0.178}}} &  {\textbf{0.267}}
&\multicolumn{1}{|c}{ {\textbf{0.171}}} &  {\underline{0.270}}
&\multicolumn{1}{|c}{0.182} & 0.272
&\multicolumn{1}{|c}{0.178} & 0.270
&\multicolumn{1}{|c}{0.216} & 0.318
&\multicolumn{1}{|c}{0.193} & 0.304
&\multicolumn{1}{|c}{0.196} & 0.309
&\multicolumn{1}{|c}{0.225} & 0.319
&\multicolumn{1}{|c}{0.223} & 0.302
&\multicolumn{1}{|c}{0.270} & 0.288

\\
\midrule
\multicolumn{2}{c}{$1^{st}$ } 
&\multicolumn{2}{|c}{ {\textbf{40}} }
&\multicolumn{2}{|c}{ {\underline{24}}} 
&\multicolumn{2}{|c}{2} 
&\multicolumn{2}{|c}{0} 
&\multicolumn{2}{|c}{0} 
&\multicolumn{2}{|c}{0} 
&\multicolumn{2}{|c}{0} 
&\multicolumn{2}{|c}{0} 
&\multicolumn{2}{|c}{1} 
&\multicolumn{2}{|c}{2}

\\
\midrule
\end{tabular}

    \caption{Full results of the multivariate long-term forecasting result comparison. The input sequence
length is set to 96 for all baselines and the prediction lengths $F \in \{96, 192, 336, 720\}$. Avg means
the average results from all four prediction lengths.
\label{tab:full}}
\end{table*}

\subsection{Frequency K-top Selection}

In real-world scenarios, a vast number of multivariate time series exhibit complex and diverse frequency components. Moreover, among the numerous frequency constituents within these time series, not all contribute meaningfully to the representation. These sequences commonly contain noise that reduces the signal-to-noise ratio of the time series, thereby leading to suboptimal performance.

To address this, we designed Frequency K-top Selection (FreK) module, which reduces noise through multi-scale principal frequency selection while comprehensively capturing temporal representation from the time series.

\subsubsection{Frequency Band Selection}

The FreK module first employs its Frequency Band Selection (FBS) block to screen primary components of time series through energy-distribution-based filtering. Since multivariate time series exhibit complex energy distributions that are difficult to extract directly, inspired by Theorem~\ref{Parseval}, we transform the time series into the spectral domain, initiating processing from the distribution of frequency components. Furthermore, to mitigate noise interference in time series and enhance the signal-to-noise ratio, we rank frequency bands in descending order of spectral energy and select the top-K bands as primary constituents of the time series. These selected bands are then inversely transformed back to the temporal domain to reconstruct the time series. As demonstrated in Theorem~\ref{Distribution}, controlling the energy distribution threshold enables reconstruction of time series that optimally approximates the noise-free sequence. Here, the reconstructed series $\tilde{x}(t)$ comes as followed:
\begin{equation}
    \tilde{x}(t) = rFFT(TopK(FFT(x(t)))
\end{equation}
where $K$ is the minimum value conducted as followed:
\begin{equation}
   \frac{\sum_{i=1}^K{X[i]^2}}{\sum_{i=1}^{L/2+1}{X[i]^2}} >\delta
\end{equation}
\begin{equation}
    \{X(k)\}_{k=1}^{L/2+1} = sorted[FFT(x(t))]
\end{equation}
where $sorted[\cdot]$ donates sorting by magnitude in descending order, $\delta$ donates the threshold of energy percentage. 

At this stage, $\tilde{x}(t)$ consists predominantly of channel-independent temporal information with lower noise. Subsequently, Frek performs Adaptive Embedding(AE) of x along the temporal dimension and employs KAN for representation learning of intrinsic information. This process can be represented by the following formula:
\begin{equation}
    E_1 = KAN(AE(\tilde{x}(t)))
\end{equation}
where $AE(\cdot )$ donates the adaptive embedding, $E_1$ donates the temporal representation by FreK. 

\subsubsection{Adaptive Embedding}
 In contemporary state-of-the-art time series forecasting models, the integration of adaptive modules into embeddings is frequently addressed. We also introduce an adaptive parameter $P \in \mathbb{R}^{D}$ to improve prediction efficacy, here $D$ donates the dimension of embedding space. However, unlike these approaches ~\cite{wang2024timexer}, the adaptive parameter in adaptive embedding serves to learn distinct characteristics unique to each dataset. The usage of $P$ with one input series $x_i \in \mathbb{R}^{L \times C}$ is as follows:
 \begin{equation}
     E_i^j=concat([P,Linear(x_i^j)])
 \end{equation} where j donates the index of variate.
 Thus, the whole embedding is expressed as follows:
 \begin{equation}
     E_i = AE(x_i) = [E_i^1,E_i^2,\cdots,E_i^{d_{model}}]
 \end{equation}

\subsubsection{Group-Rational KAN}

Compared to traditional MLPs, KAN replaces fixed activate functions with learnable univariate functions, allowing complex nonlinear relationships to be modeled with fewer parameters and greater interpretability. In our methodology, we employ Group-Rational KANs~\cite{yang2025kolmogorovarnold} to learn representations of temporal components. The rational base functions are constructed by Q(x) and P(x) of order m,n.
\begin{equation}
    \phi (x)=wF(x)=w \frac{P(x)}{Q(x)} = w \frac{\sum_{i=0}^m a_ix^i}{\sum_{i=0}^m b_ix^i}
\end{equation} 
where $a_i$ and $b_i$ are coefficient of the rational function and $w$ is the scaling factor. 

To integrate rational functions as base functions within KANs while mitigating the instability caused by poles, which occurs when Q(x)=0, Group-KAN employs a modified formulation of the standard rational function.
\begin{equation}
    F(x) = \frac{a_0+a_1x+...+a_mx^m}{1+|b_1x+...+b_mx^m|}
\end{equation}

Thus, Group-Rational KAN incorporates rational functions and constructs its processing architecture through group seperation and sharing base function within group. For an input variable $X \in \mathbb{R}^{d_{in}}$, let i denotes its channel index. With $g$ groups, each group in GR-KAN contains $d_g=d_{in}/g$ channels, where $\lfloor i/d 
g\rfloor$ represents the group index. The operation of GR-KAN on x can be expressed as:
\begin{equation}
    GR\text{-}KAN(\mathbf{x})  = \phi \  \circ \ x=WF(x)
\end{equation}
To simplify it, we express it in matrix form as the product of a weight matrix $W \in \mathbb{R}^{d_{in}\times d_{out}}$ and a rational function F:
\begin{equation}
W= 
\begin{bmatrix}
w_{1,1}       & \cdots & w_{1,d_{\text{in}}} \\
\vdots        & \ddots & \vdots             \\
w_{d_{\text{out}},1} & \cdots & w_{d_{\text{out}},d_{\text{in}}}
\end{bmatrix}
\end{equation}

\begin{equation}   
F(x)=\begin{bmatrix}
F_{\lfloor 1/d_g \rfloor}(x_1) &
\cdots 
 &F_{\lfloor d_{\text{in}}/d_g \rfloor}(x_{d_{\text{in}}})
\end{bmatrix}^T
\end{equation}

In our implementation of Rational KAN, we simply prefix the rational function to a linear layer as a unit of KAN. And the KAN used in our work is consist of two units.
\begin{equation}
    KAN_i(x)=linear(F(x))
\end{equation}
where i donates the layer index in our KAN.

\subsection{Time Stamp Embedding}
Additionally, we introduce linear embeddings for timestamps. In the real world, physical quantities closely associated with time series, such as mechanical load and electricity consumption, exhibit daily, monthly, yearly, and other levels of periodicity along the temporal dimension. By aligning timestamp information with the latent representations learned by the model, we can further enhance the model’s ability to understand time series data. While existing approaches~\cite{wang2024timexer} incorporate timestamp information to boost model performance, they neglect the critical synchronization of temporal markers with multi-scale sequence patterns. Our methodology resolves this through time stamp down sampling, where temporal embeddings are progressively coarsened to maintain alignment with corresponding resolution levels in the input sequence hierarchy.

\subsection{Feature Mixing}

After specifically learning temporal information from time series at different scales, we need to organically integrate the feature representations learned by the model. Here, we refer to the widely adopted feedforward network. In contrast, we incorporate timestamp information from the time series and replace the MLP with a KAN.

As such, the feature mixing module can be represented by the following formula:
\begin{equation}
    FM(E_1,E_s)=E_1+KAN([E_1,E_s])
\end{equation}
where $E_s$ donates the linear embedding of time stamps.

For the fused multiscale data, we employ average aggregation followed by a simple linear projection layer to generate the predicted output $\tilde{y}(t)$:
\begin{equation}
    \tilde{y}(t) = linear(FM_{avg})
\end{equation}
where $FM_{avg}$ donates the mean FM output on different input scales.

\subsection{Loss Function}

Incorporating frequency domain alignment terms into loss functions is not novel. However, unlike previous work ~\cite{wang2025fredf}, our approach enforces alignment exclusively on the dominant frequencies of the data. While this method reduces fine-grained fitting precision, we maintain that frequency-domain signals primarily serve as coarse-grained representations for capturing macro-level trend shifts. Intuitively, fine-grained information modeling can be sufficiently handled by the MSE loss function alone. The specific formulation is shown as follows.
\begin{equation}
    \mathcal{L}_{F} = \frac{1}{K} \sum_i^K||\mathcal{F}\{\tilde{y}(t)\}_i-\mathcal{F}\{y(t)\}_i||
\end{equation}

By combining the hybrid loss $L_{F}$ with the MSE loss, we arrive at our final loss function as follows:
\begin{equation}
\mathcal{L} = \alpha \mathcal{L}_F+(1-\alpha) \mathcal{L}_{\text{MSE}}
\end{equation}
where $\alpha$ is a hyperparameter, $\tilde{y}(t)$ donates the prediction of KFS, K donates the index of top-K frequency prediction data with the highest amplitudes. Unlike FreK, the K here is set to a fixed value of 32. This loss function accounts for both temporal discrepancies and introduces alignment of the principal frequencies in the time series.

\section{Experiments}
\subsubsection{Datasets}

We conducted long-term forecasting experiments on six real-world datasets: ETT-Series~\cite{haoyietal-informer-2021}, Electricity~\cite{electricityloaddiagrams20112014_321} and Weather~\cite{haoyietal-informer-2021}. Following established protocols from previous studies ~\cite{wu2023timesnet,wang2024tssurvey}, we split the datasets of the ETT series into training, validation, and test sets according to a 6: 2: 2 ratio. For the remaining datasets, the ratio is 7:1:2.

\subsubsection{Baselines}

We carefully selected representative models as baselines in field of time series forecasting, including: 
1)Transformer-based models: TimeXer~\cite{wang2024timexer}, PatchTST~\cite{nie2023a}, iTransformer~\cite{liu2024itransformer}.
2)CNN-based models: TimesNet~\cite{wu2023timesnet}, MICN~\cite{wang2023micn}.
3)MLP-based models: TimeMixer~\cite{wang2024timemixer}, DLinear~\cite{Zeng2022AreTE}.
4)Frequency-based models:  FiLM~\cite{zhou2022film}.
 And a time series foundation model Time-FFM~\cite{liu2024timeffm}.
\subsubsection{Experimental Settings}
To ensure fair comparisons, we adopt the same look-back window length $T = 96$ and the same prediction length $F = \{96, 192, 336, 720\}$. We use Mean Square Error (MSE) and Mean Absolute Error (MAE) metrics to evaluate the
performance of each method.

\subsection{Main Results}
Comprehensive forecasting results are shown in Table~\ref{tab:full}, the best results are highlighted in  {\textbf{Bold}} and the second-best are \underline{underlined}. Lower MSE/MAE values indicate higher prediction accuracy. We observe that KFS demonstrates exceptional performance in all datasets except for the ECL dataset. TimeXer achieves optimal results on this particular dataset, primarily because its cross-attention mechanism provides a strong ability of learning inter-channel relationships. This architecture enables TimeXer to better model channel dependencies, an advantage particularly pronounced in high-dimensional datasets like Electricity.

Furthermore, both TimeXer and KFS consistently perform well in long-term forecasting tasks, demonstrating the models' strong generalization capabilities and KFS's well-designed framework. Compared with other SOTA models, KFS introduces an innovative frequency-domain processing method for time series, extending multivariate forecasting frameworks in a new form. By leveraging the characteristics of multi-scale time series frameworks and skillfully integrating specialized frequency-domain processing with diverse feature representations, KFS achieves outstanding performance in multiple time series forecasting tasks.

\subsection{Model Analysis}
\subsubsection{Ablation Study}

To investigate the effectiveness of each component of KFS, we perform detailed ablation of each possible design on weather and ETTh2 datasets. As show in Tab~\ref{tab:ablation}, we have following observations.

\begin{table}[!h]
    \centering
    \small
\setlength{\tabcolsep}{2pt}

\begin{tabular}{*{10}{c}} 
\toprule
\multicolumn{2}{c}{Models}
 & \multicolumn{2}{c}{KFS} & \multicolumn{2}{c}{KAN$\rightarrow$MLP} & \multicolumn{2}{c}{w/o Stamp} & \multicolumn{2}{c}{w/o AE}  \\
\cmidrule(lr){3-4} \cmidrule(lr){5-6} \cmidrule(lr){7-8}   \cmidrule(lr){9-10} 

\multicolumn{2}{c}{Metric} &MSE &\multicolumn{1}{c|}{MAE} &MSE &\multicolumn{1}{c|}{MAE} &MSE &\multicolumn{1}{c|}{MAE} &MSE &\multicolumn{1}{c}{MAE}   \\
\midrule

\multirow{5}{*}{\begin{turn}{90}weather\end{turn}} 
&\multicolumn{1}{|c}{96} 
&\multicolumn{1}{|c}{0.159} & 0.205
&\multicolumn{1}{|c}{0.161} & 0.205 
&\multicolumn{1}{|c}{0.163} & 0.208
&\multicolumn{1}{|c}{0.163} & 0.209
\\
&\multicolumn{1}{|c}{192} 
&\multicolumn{1}{|c}{0.207} & 0.249
&\multicolumn{1}{|c}{0.208} & 0.249
&\multicolumn{1}{|c}{0.211} & 0.251
&\multicolumn{1}{|c}{0.211} & 0.252
\\
&\multicolumn{1}{|c}{336} 
&\multicolumn{1}{|c}{0.262} & 0.288
&\multicolumn{1}{|c}{0.264} & 0.289
&\multicolumn{1}{|c}{0.262} & 0.289
&\multicolumn{1}{|c}{0.262} & 0.288
\\
&\multicolumn{1}{|c}{720} 
&\multicolumn{1}{|c}{0.345} & 0.342
&\multicolumn{1}{|c}{0.342} & 0.340
&\multicolumn{1}{|c}{0.344} & 0.343
&\multicolumn{1}{|c}{0.347} & 0.344
\\

\cmidrule(r){2-10}
&\multicolumn{1}{|c}{Avg} 
&\multicolumn{1}{|c}{ {\textbf{0.243}}} &  {\textbf{0.271}}
&\multicolumn{1}{|c}{0.244} & 0.271
&\multicolumn{1}{|c}{0.245} & 0.272
&\multicolumn{1}{|c}{0.245} & 0.273

\\

\midrule
\multirow{5}{*}{\begin{turn}{90}ETTh2\end{turn}} 
&\multicolumn{1}{|c}{96} 
&\multicolumn{1}{|c}{0.280} & 0.334
&\multicolumn{1}{|c}{0.284} & 0.337
&\multicolumn{1}{|c}{0.282} & 0.335
&\multicolumn{1}{|c}{0.279} & 0.334
\\
&\multicolumn{1}{|c}{192} 
&\multicolumn{1}{|c}{0.362} & 0.387
&\multicolumn{1}{|c}{0.366} & 0.388
&\multicolumn{1}{|c}{0.365} & 0.386
&\multicolumn{1}{|c}{0.364} & 0.386
\\
&\multicolumn{1}{|c}{336} 
&\multicolumn{1}{|c}{0.406} & 0.421
&\multicolumn{1}{|c}{0.419} & 0.426
&\multicolumn{1}{|c}{0.410} & 0.422
&\multicolumn{1}{|c}{0.414} & 0.425
\\
&\multicolumn{1}{|c}{720} 
&\multicolumn{1}{|c}{0.423} & 0.435
&\multicolumn{1}{|c}{0.435} & 0.443
&\multicolumn{1}{|c}{0.431} & 0.444
&\multicolumn{1}{|c}{0.431} & 0.440
\\

\cmidrule(r){2-10}
&\multicolumn{1}{|c}{Avg} 
&\multicolumn{1}{|c}{ {\textbf{0.367}}} &  {\textbf{0.394}}
&\multicolumn{1}{|c}{0.373} & 0.399
&\multicolumn{1}{|c}{0.372} & 0.396
&\multicolumn{1}{|c}{0.372} & 0.396

\\

\midrule
\end{tabular}

\caption{Rsults of Ablation Study on weather and ETTh2.
\label{tab:ablation}}
\end{table}
\begin{figure}[t]
\centering
\includegraphics[width=1\columnwidth]{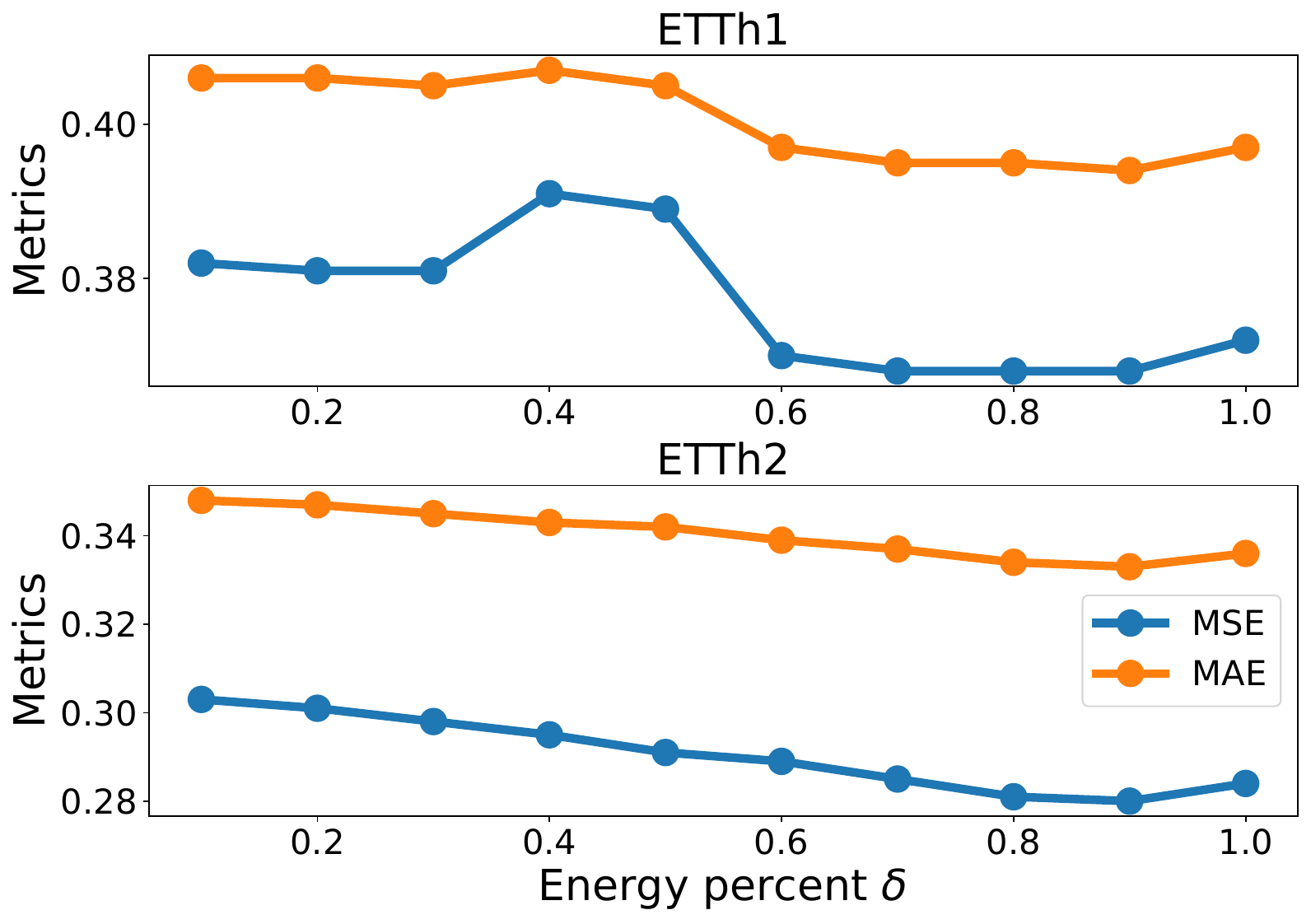} 

\caption{The impact of $\delta$ on metrics. This experiment is conducted on ETTh1 and ETTh2 datasets with look-back window 96 and prediction length 96.
\label{fig:Percent}}
\end{figure}
\begin{table}[!h]
    \centering
    \small
\setlength{\tabcolsep}{3pt}
\begin{tabular}{*{8}{c}} 
\toprule
\multicolumn{2}{c}{Methods}
 & \multicolumn{2}{c}{Top-K(Ours)} & \multicolumn{2}{c}{Avg Filter} & \multicolumn{2}{c}{Gaussian Filter}   \\
\cmidrule(lr){3-4} \cmidrule(lr){5-6} \cmidrule(lr){7-8}  

\multicolumn{2}{c}{Metric} &MSE &\multicolumn{1}{c|}{MAE} &MSE &\multicolumn{1}{c|}{MAE} &MSE &\multicolumn{1}{c}{MAE}   \\
\midrule

\multirow{5}{*}{\begin{turn}{90}weather\end{turn}} 
&\multicolumn{1}{|c}{96} 
&\multicolumn{1}{|c}{0.159} & 0.205
&\multicolumn{1}{|c}{0.161} & 0.208
&\multicolumn{1}{|c}{0.160} & 0.206
\\
&\multicolumn{1}{|c}{192} 
&\multicolumn{1}{|c}{0.207} & 0.249
&\multicolumn{1}{|c}{0.214} & 0.255
&\multicolumn{1}{|c}{0.211} & 0.252
\\
&\multicolumn{1}{|c}{336} 
&\multicolumn{1}{|c}{0.262} & 0.288
&\multicolumn{1}{|c}{0.272} & 0.294
&\multicolumn{1}{|c}{0.265} & 0.289
\\
&\multicolumn{1}{|c}{720} 
&\multicolumn{1}{|c}{0.345} & 0.342
&\multicolumn{1}{|c}{0.346} & 0.342
&\multicolumn{1}{|c}{0.345} & 0.343
\\

\cmidrule(r){2-8}
&\multicolumn{1}{|c}{Avg} 
&\multicolumn{1}{|c}{ {\textbf{0.243}}} &  {\textbf{0.271}}
&\multicolumn{1}{|c}{0.248} & 0.274
&\multicolumn{1}{|c}{0.245} & 0.272

\\

\midrule
\multirow{5}{*}{\begin{turn}{90}ETTh2\end{turn}} 
&\multicolumn{1}{|c}{96} 
&\multicolumn{1}{|c}{0.280} & 0.334
&\multicolumn{1}{|c}{0.292} & 0.345
&\multicolumn{1}{|c}{0.283} & 0.336
\\
&\multicolumn{1}{|c}{192} 
&\multicolumn{1}{|c}{0.362} & 0.387
&\multicolumn{1}{|c}{0.380} & 0.399
&\multicolumn{1}{|c}{0.364} & 0.388
\\
&\multicolumn{1}{|c}{336} 
&\multicolumn{1}{|c}{0.406} & 0.421
&\multicolumn{1}{|c}{0.423} & 0.431
&\multicolumn{1}{|c}{0.407} & 0.423
\\
&\multicolumn{1}{|c}{720} 
&\multicolumn{1}{|c}{0.423} & 0.435
&\multicolumn{1}{|c}{0.437} & 0.448
&\multicolumn{1}{|c}{0.433} & 0.443
\\

\cmidrule(r){2-8}
&\multicolumn{1}{|c}{Avg} 
&\multicolumn{1}{|c}{ {\textbf{0.367}}} &  {\textbf{0.394}}
&\multicolumn{1}{|c}{0.383} & 0.405
&\multicolumn{1}{|c}{0.372} & 0.397

\\

\midrule
\end{tabular}

    \caption{Rsults of Filter Study On weather and ETTh2 dataset.
\label{tab:Filter}}
\end{table}
For KAN, we substituted it into a standard MLP with matched parameterization. The consequent deterioration in error metrics substantiates that KFS's implementation of the KAN delivers substantially stronger representation learning than conventional MLPs. This evidence validates the functional superiority of rational basis functions for TSF.

For Time Stamp part \textbf{(w/o Stamp)}, We replaced the time stamp embedding with a zero matrix of identical dimensions. We observed performance degradation on both datasets when removing the TimeStamp component. Notably, the performance decline was more pronounced on ETTh2 that is an electricity equipment load dataset exhibiting stronger temporal periodicity compared to the Weather dataset. This outcome empirically validates the simple yet effective design of our TimeStamp embedding methodology.

For Embedding method \textbf{(w/o AE)}, We substituted the learnable parameter $P$ in the Adaptive Embedding with a fixed zero matrix of identical shape and dimensions.

\begin{figure}[t]
\centering
\includegraphics[width=1\columnwidth]{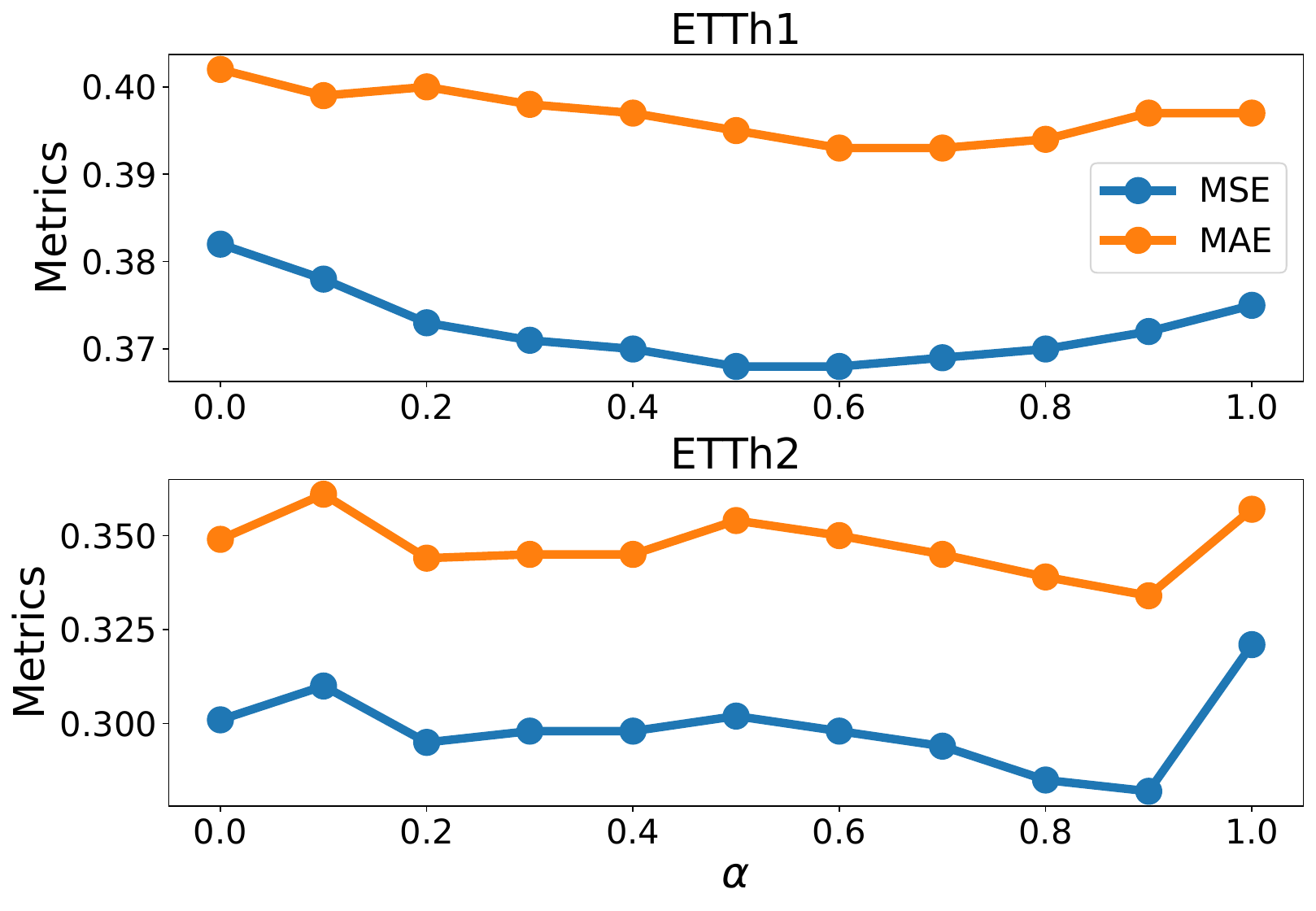} 
\caption{The impact of $\alpha$ on metrics. This experiment is conducted on ETTh1 dataset with prediction length 96.}
\label{fig:LOSS}
\end{figure}

For FreK, we conducted two experiments to investigate the effect of Top-K Selection. In one experiment, we evaluated the impact of $\delta$ on model performance using the ETTh1 and ETTh2 datasets. In another experiment, we examined how alternative filtering methods (e.g. mean filtering and Gaussian filtering) affect model effectiveness on both the ETTh2 and Weather datasets. The results of these two experiments are presented in Figure~\ref{fig:Percent} and Table~\ref{tab:Filter}, respectively. From the results, we observe that as $\delta$ increases, model metrics generally reach their minimum at $\delta=0.9$, indicating that our energy-threshold-based frequency selection strategy improves the performance of the model. Furthermore, compared to alternative filtering methods, our approach achieves superior results, validating the effectiveness of the Top-K selection strategy in mitigating noise interference.

For the loss term, we conducted a dedicated experiment on the ETTh1 dataset to investigate the impact of $\alpha$ on model performance, with experimental results presented in Figure~\ref{fig:LOSS}. The experimental results demonstrate appropriate calibration of $\alpha$ substantially enhances model capabilities, experimentally validating the efficacy of our proposed loss function combination.

\begin{table}[]
    \centering
    \small
    \begin{tabular}{c|c|c|c}
    \toprule
        Model & Memory & Step Time & FLOPs \\
        \midrule
        PatchTST & 807 MB & 70 ms&51.28 GB \\
        \midrule
        FEDformer&379 MB &70 ms &5.28 GB \\
        \midrule
        TimesNet&1227 MB&50 ms&115.85 GB\\
        \midrule
        TimeMixer& {\underline{132 MB}}& {\textbf{13 ms}}& {\textbf{0.62 GB}}\\
        \midrule
        KFS&  {\textbf{116 MB}}&  {\underline{21 ms}}& {\underline{1.66 GB}}\\
        \midrule
    \end{tabular}
    \caption{A comparison of used Memory, Training Time per step and FLOPs between KFS and other 4 models. To ensure a fair comparison, we fix the prediction length $F = 96$ and the input length $T = 96$, and set the input batch size to 32. }
    \label{tab:Efficiency}
\end{table}

\subsubsection{Efficiency Analysis}
We conducted a comprehensive comparison of training time, used memory and FLOPs in various baseline models in the Weather dataset, using official model configurations and identical batch size. The results are shown in Table~\ref{tab:Efficiency}. It is clear that our KFS demonstrates significant advantages in memory cost in all models. Moreover, the efficiency of KFS outperforms other Transformer-based and CNN-based models. Furthermore, the training time and FLOPs reveals that despite KFS's incorporation of FFT operations, which increase computational complexity, the overall training efficiency remains competitive.

\section{Conclusion}

In this paper, we propose the KAN-based long term Time series forecasting(KFS) framework  to address spectral noise entanglement in complex time series.  Comprehensive experiments demonstrate that KFS achieves state-of-the-art performance in long-term forecasting tasks across diverse datasets, showcasing superior efficiency and effectiveness.

\bibliography{aaai2026}

\begin{thebibliography}{28}
\providecommand{\natexlab}[1]{#1}

\bibitem[{Aghaei(2024)}]{aghaei2024rkanrationalkolmogorovarnoldnetworks}
Aghaei, A.~A. 2024.
\newblock rKAN: Rational Kolmogorov-Arnold Networks.
\newblock arXiv:2406.14495.

\bibitem[{Bodner et~al.(2025)Bodner, Tepsich, Spolski, and
  Pourteau}]{bodner2025convolutionalkolmogorovarnoldnetworks}
Bodner, A.~D.; Tepsich, A.~S.; Spolski, J.~N.; and Pourteau, S. 2025.
\newblock Convolutional Kolmogorov-Arnold Networks.
\newblock arXiv:2406.13155.

\bibitem[{donghao and wang xue(2024)}]{donghao2024moderntcn}
donghao, L.; and wang xue. 2024.
\newblock Modern{TCN}: A Modern Pure Convolution Structure for General Time
  Series Analysis.
\newblock In \emph{The Twelfth International Conference on Learning
  Representations}.

\bibitem[{Huang, Chen, and Qiao(2024)}]{huang2024generative}
Huang, H.; Chen, M.; and Qiao, X. 2024.
\newblock Generative Learning for Financial Time Series with Irregular and
  Scale-Invariant Patterns.
\newblock In \emph{The Twelfth International Conference on Learning
  Representations}.

\bibitem[{Huang et~al.(2025)Huang, Zhao, Li, and BAI}]{huang2025timekan}
Huang, S.; Zhao, Z.; Li, C.; and BAI, L. 2025.
\newblock Time{KAN}: {KAN}-based Frequency Decomposition Learning Architecture
  for Long-term Time Series Forecasting.
\newblock In \emph{The Thirteenth International Conference on Learning
  Representations}.

\bibitem[{Jiang and Luo(2022)}]{JIANG2022117921}
Jiang, W.; and Luo, J. 2022.
\newblock Graph neural network for traffic forecasting: A survey.
\newblock \emph{Expert Systems with Applications}, 207: 117921.

\bibitem[{Lam et~al.(2023)Lam, Sanchez-Gonzalez, Willson, Wirnsberger,
  Fortunato, Alet, Ravuri, Ewalds, Eaton-Rosen, Hu, Merose, Hoyer, Holland,
  Vinyals, Stott, Pritzel, Mohamed, and
  Battaglia}]{doi:10.1126/science.adi2336}
Lam, R.; Sanchez-Gonzalez, A.; Willson, M.; Wirnsberger, P.; Fortunato, M.;
  Alet, F.; Ravuri, S.; Ewalds, T.; Eaton-Rosen, Z.; Hu, W.; Merose, A.; Hoyer,
  S.; Holland, G.; Vinyals, O.; Stott, J.; Pritzel, A.; Mohamed, S.; and
  Battaglia, P. 2023.
\newblock Learning skillful medium-range global weather forecasting.
\newblock \emph{Science}, 382(6677): 1416--1421.

\bibitem[{Li(2024)}]{li2024kolmogorovarnoldnetworksradialbasis}
Li, Z. 2024.
\newblock Kolmogorov-Arnold Networks are Radial Basis Function Networks.
\newblock arXiv:2405.06721.

\bibitem[{Lin et~al.(2024)Lin, Lin, Wu, Chen, and Yang}]{lin2024sparsetsf}
Lin, S.; Lin, W.; Wu, W.; Chen, H.; and Yang, J. 2024.
\newblock Sparse{TSF}: Modeling Long-term Time Series Forecasting with *1k*
  Parameters.
\newblock In \emph{Forty-first International Conference on Machine Learning}.

\bibitem[{Liu et~al.(2024{\natexlab{a}})Liu, Liu, Liu, Wen, and
  Liang}]{liu2024timeffm}
Liu, Q.; Liu, X.; Liu, C.; Wen, Q.; and Liang, Y. 2024{\natexlab{a}}.
\newblock Time-{FFM}: Towards {LM}-Empowered Federated Foundation Model for
  Time Series Forecasting.
\newblock In \emph{The Thirty-eighth Annual Conference on Neural Information
  Processing Systems}.

\bibitem[{Liu et~al.(2024{\natexlab{b}})Liu, Hu, Zhang, Wu, Wang, Ma, and
  Long}]{liu2024itransformer}
Liu, Y.; Hu, T.; Zhang, H.; Wu, H.; Wang, S.; Ma, L.; and Long, M.
  2024{\natexlab{b}}.
\newblock iTransformer: Inverted Transformers Are Effective for Time Series
  Forecasting.
\newblock In \emph{The Twelfth International Conference on Learning
  Representations}.

\bibitem[{Liu et~al.(2025)Liu, Wang, Vaidya, Ruehle, Halverson, Soljacic, Hou,
  and Tegmark}]{liu2025kan}
Liu, Z.; Wang, Y.; Vaidya, S.; Ruehle, F.; Halverson, J.; Soljacic, M.; Hou,
  T.~Y.; and Tegmark, M. 2025.
\newblock {KAN}: Kolmogorov{\textendash}Arnold Networks.
\newblock In \emph{The Thirteenth International Conference on Learning
  Representations}.

\bibitem[{Nie et~al.(2023)Nie, Nguyen, Sinthong, and Kalagnanam}]{nie2023a}
Nie, Y.; Nguyen, N.~H.; Sinthong, P.; and Kalagnanam, J. 2023.
\newblock A Time Series is Worth 64 Words: Long-term Forecasting with
  Transformers.
\newblock In \emph{The Eleventh International Conference on Learning
  Representations}.

\bibitem[{Trindade(2015)}]{electricityloaddiagrams20112014_321}
Trindade, A. 2015.
\newblock {ElectricityLoadDiagrams20112014}.
\newblock UCI Machine Learning Repository.
\newblock {DOI}: https://doi.org/10.24432/C58C86.

\bibitem[{Wang et~al.(2025)Wang, Pan, Shen, Chen, Yang, Yang, Zhang, Liu, Li,
  and Tao}]{wang2025fredf}
Wang, H.; Pan, L.; Shen, Y.; Chen, Z.; Yang, D.; Yang, Y.; Zhang, S.; Liu, X.;
  Li, H.; and Tao, D. 2025.
\newblock Fre{DF}: Learning to Forecast in the Frequency Domain.
\newblock In \emph{The Thirteenth International Conference on Learning
  Representations}.

\bibitem[{Wang et~al.(2023)Wang, Peng, Huang, Wang, Chen, and
  Xiao}]{wang2023micn}
Wang, H.; Peng, J.; Huang, F.; Wang, J.; Chen, J.; and Xiao, Y. 2023.
\newblock {MICN}: Multi-scale Local and Global Context Modeling for Long-term
  Series Forecasting.
\newblock In \emph{The Eleventh International Conference on Learning
  Representations}.

\bibitem[{Wang et~al.(2024{\natexlab{a}})Wang, Wu, Shi, Hu, Luo, Ma, Zhang, and
  ZHOU}]{wang2024timemixer}
Wang, S.; Wu, H.; Shi, X.; Hu, T.; Luo, H.; Ma, L.; Zhang, J.~Y.; and ZHOU, J.
  2024{\natexlab{a}}.
\newblock TimeMixer: Decomposable Multiscale Mixing for Time Series
  Forecasting.
\newblock In \emph{The Twelfth International Conference on Learning
  Representations}.

\bibitem[{Wang et~al.(2024{\natexlab{b}})Wang, Wu, Dong, Liu, Long, and
  Wang}]{wang2024tssurvey}
Wang, Y.; Wu, H.; Dong, J.; Liu, Y.; Long, M.; and Wang, J. 2024{\natexlab{b}}.
\newblock Deep Time Series Models: A Comprehensive Survey and Benchmark.

\bibitem[{Wang et~al.(2024{\natexlab{c}})Wang, Wu, Dong, Liu, Qiu, Zhang, Wang,
  and Long}]{wang2024timexer}
Wang, Y.; Wu, H.; Dong, J.; Liu, Y.; Qiu, Y.; Zhang, H.; Wang, J.; and Long, M.
  2024{\natexlab{c}}.
\newblock Timexer: Empowering transformers for time series forecasting with
  exogenous variables.
\newblock \emph{Advances in Neural Information Processing Systems}.

\bibitem[{Wu et~al.(2023)Wu, Hu, Liu, Zhou, Wang, and Long}]{wu2023timesnet}
Wu, H.; Hu, T.; Liu, Y.; Zhou, H.; Wang, J.; and Long, M. 2023.
\newblock TimesNet: Temporal 2D-Variation Modeling for General Time Series
  Analysis.
\newblock In \emph{International Conference on Learning Representations}.

\bibitem[{Wu et~al.(2021)Wu, Xu, Wang, and Long}]{NEURIPS2021_bcc0d400}
Wu, H.; Xu, J.; Wang, J.; and Long, M. 2021.
\newblock Autoformer: Decomposition Transformers with Auto-Correlation for
  Long-Term Series Forecasting.
\newblock In Ranzato, M.; Beygelzimer, A.; Dauphin, Y.; Liang, P.; and Vaughan,
  J.~W., eds., \emph{Advances in Neural Information Processing Systems},
  volume~34, 22419--22430. Curran Associates, Inc.

\bibitem[{Xingyi~Yang(2025)}]{yang2025kolmogorovarnold}
Xingyi~Yang, X.~W. 2025.
\newblock Kolmogorov-Arnold Transformer.
\newblock In \emph{The Thirteenth International Conference on Learning
  Representations}.

\bibitem[{Xu, Zeng, and Xu(2024)}]{xu2024fits}
Xu, Z.; Zeng, A.; and Xu, Q. 2024.
\newblock {FITS}: Modeling Time Series with \$10k\$ Parameters.
\newblock In \emph{The Twelfth International Conference on Learning
  Representations}.

\bibitem[{Zeng et~al.(2023)Zeng, Chen, Zhang, and Xu}]{Zeng2022AreTE}
Zeng, A.; Chen, M.; Zhang, L.; and Xu, Q. 2023.
\newblock Are Transformers Effective for Time Series Forecasting?

\bibitem[{Zhang and Yan(2023)}]{zhang2023crossformer}
Zhang, Y.; and Yan, J. 2023.
\newblock Crossformer: Transformer Utilizing Cross-Dimension Dependency for
  Multivariate Time Series Forecasting.
\newblock In \emph{International Conference on Learning Representations}.

\bibitem[{Zhou et~al.(2021)Zhou, Zhang, Peng, Zhang, Li, Xiong, and
  Zhang}]{haoyietal-informer-2021}
Zhou, H.; Zhang, S.; Peng, J.; Zhang, S.; Li, J.; Xiong, H.; and Zhang, W.
  2021.
\newblock Informer: Beyond Efficient Transformer for Long Sequence Time-Series
  Forecasting.
\newblock In \emph{The Thirty-Fifth {AAAI} Conference on Artificial
  Intelligence, {AAAI} 2021, Virtual Conference}, volume~35, 11106--11115.
  {AAAI} Press.

\bibitem[{Zhou et~al.(2022{\natexlab{a}})Zhou, Ma, Wen, Wang, Sun, and
  Jin}]{zhou2022fedformer}
Zhou, T.; Ma, Z.; Wen, Q.; Wang, X.; Sun, L.; and Jin, R. 2022{\natexlab{a}}.
\newblock {FEDformer}: Frequency enhanced decomposed transformer for long-term
  series forecasting.
\newblock In \emph{Proc. 39th International Conference on Machine Learning
  (ICML 2022)}.

\bibitem[{Zhou et~al.(2022{\natexlab{b}})Zhou, Ma, xue wang, Wen, Sun, Yao,
  Yin, and Jin}]{zhou2022film}
Zhou, T.; Ma, Z.; xue wang; Wen, Q.; Sun, L.; Yao, T.; Yin, W.; and Jin, R.
  2022{\natexlab{b}}.
\newblock Fi{LM}: Frequency improved Legendre Memory Model for Long-term Time
  Series Forecasting.
\newblock In Oh, A.~H.; Agarwal, A.; Belgrave, D.; and Cho, K., eds.,
  \emph{Advances in Neural Information Processing Systems}.

\end{thebibliography}

\section{Appendix}
\subsection{Proof of Theory 2}
\begin{theorem}\textbf{Theory 2}
Let observed time series \( y = y_0 + n \), where \( n \sim \mathcal{N}(0, \sigma^2 I) \), and $y_0$ donates original times series. After DFT, there exist \( K \in \mathbb{N}^+ \) and \( \epsilon > 0 \) such that the sparse reconstruction \( \tilde{y} \) from the top-$K$ frequencies of \( Y \) satisfies:
\begin{equation}
\|\tilde{y} - y_0\|_2 < \epsilon.
\end{equation}
\label{Distribution}
\end{theorem}

$Proof.$ To prove Theorem 2, it suffices to demonstrate that $P(\|\tilde{y} - y_0\|_2 > \epsilon)$ has an upper bound $\mu<1$, which reduces to verifying the following inequality:

\[
\mathbb{P}\left( \| \tilde{y} - y_0 \|_2^2 > \epsilon \right) \leq \frac{C \sigma^2 K + \sum_{j>K} |Y_0^{(j)}|^2}{\epsilon} <1,
\]
where \( Y_0^{(j)} \) is the \( j \)-th largest DFT coefficient of \( y_0 \), $C > 0$ is a constant value.

The reconstruction error decomposes into two parts:
\[
\| \tilde{y} - y_0 \|_2^2 = \underbrace{\sum_{j>K} |Y_0^{(j)}|^2}_{\text{Signal Energy Loss}} + \underbrace{\sum_{j=1}^{K} |N^{(j)}|^2}_{\text{Retained Noise Energy}}.
\]

\begin{enumerate}
    \item \textbf{Signal energy loss}: 
    By Parseval's theorem, \( \sum_{j>K} |Y_0^{(j)}|^2 \) is the energy discarded from the true signal. 
    Since \( y_0 \) has finite energy (\( \|y_0\|_2^2 < \infty \)), \( \sum_{j>K} |Y_0^{(j)}|^2 \to 0 \) as \( K \to \infty \).

    \item \textbf{Retained noise energy}: 
    The coefficients \( |N^{(j)}|^2 \) follow a chi-square distribution \( \chi^2_{2K} \) (real and imaginary parts of complex Gaussian are independent). 
    By Markov's inequality:
    \[
    \mathbb{P}\left( \sum_{j=1}^{K} |N^{(j)}|^2 > t \right) \leq \frac{\mathbb{E}\left[ \sum_{j=1}^{K} |N^{(j)}|^2 \right]}{t} = \frac{K \sigma^2}{t}.
    \]

    \item \textbf{Joint bound}: 
    For any \( \epsilon > \sum_{j>K} |Y_0^{(j)}|^2 \), set \( \epsilon_1 = \epsilon - \sum_{j>K} |Y_0^{(j)}|^2 \). Then:
    \[
    \mathbb{P}\left( \| \tilde{y} - y_0 \|_2^2 > \epsilon \right) \leq \mathbb{P}\left( \sum_{j=1}^{K} |N^{(j)}|^2 > \epsilon_1 \right) \leq \frac{K \sigma^2}{\epsilon_1}.
    \]
    Substituting \( \epsilon_1 \) gives the result with \( C = 1 \).
\end{enumerate}

Obviously, there exist at least one $K$ and one $\epsilon$ satisfying $K\sigma^2<\epsilon-\sum_{j>K}|Y_0^{(j)}|^2$, then we can get the upper bound is:
$$\frac{K\sigma^2}{\epsilon_1}=\frac{K\sigma^2}{\epsilon-\sum_{j>K}|Y_0^{(j)}|^2}<1$$

Therefore, theory 2 is proven.

\subsection{Detailed Algorithm Description}
The pseudocode of the KFS algorithm is shown in Algorithm~\ref{alg:algorithm}. The algorithm initializes the input data and parameters and performs normalization. After downsampling method, the data is then iterated through FreK block to extract multi-scale information. The time stamp is also processed by downsampling method and embedded by linear embedding method. Mixing Blocks fuse each scaled feature with its time stamp, respectively. Following this, all fused features are combined via average method at feature dimension. The prediction output is conducted by linear projection.
\begin{algorithm}[tb]
\caption{The Overall Architecture of KFS}
\label{alg:algorithm}
\textbf{Input}: look-back sequence $X \in \mathbb{R}^{L \times C}$, time stamps $T \in $.\\
\textbf{Parameter}: Energy threshold $\delta$, Loss function hyperparameter $\alpha$, Downsampling layer number n.\\
\textbf{Output}: Predictions $\hat{y}$
\begin{algorithmic}[1] 
\STATE $X$=$X^T$
\STATE $\{ X_0, X_1, \dots, X_n \}$ = Downsampling(X)
\STATE $\{ T_0, T_1, \dots, T_n \}$ = Downsampling(T)
\FOR{i in {$\{0,\dots,n\}$}}
\STATE $X_i$ = RevIn($X_i$, 'norm')
\STATE $E_1^i$ = FreK($X_i$)
\STATE $E_s^i$ = Linear($T_i$)
\STATE $FM^i$ = $E_1^i+KAN([E_1^i,E_s^i])$
\ENDFOR
\STATE $FM_{avg}$ = $\frac{1}{n+1}\sum_{i=0}^nFM_i$
\STATE $\hat{y}$ = $linear(FM_{avg})$
\STATE \textbf{return} $\hat{y}$
\end{algorithmic}
\end{algorithm}

\subsection{Details of Experiments}
\subsubsection{Detailed Dataset Descriptions}
Detailed descriptions of the data sets are shown in Table~\ref{tab:Dataset}. Dim denotes the number of channels in each dataset. Dataset Size denotes the total number of time points in (Train, Validation, Test) split, respectively. Prediction Length denotes the future time points to be predicted, and four prediction settings are included in each dataset. Frequency denotes the sampling interval of time points. Information refers to the meaning of the data.

\begin{table*}[!t]
    \centering
    \begin{tabular}{c|c|c|c|c|c}
    \toprule
        Dataset & Dim & Prediction Length & Dataset Size & Frequency & Information\\
        
        \midrule
        ETTh1, ETTh2 & 7 & $\{ 96,192,336,720\}$ & (8545, 2881, 2881) & Hourly & Electricity \\
                \midrule
        ETTm1, ETTm2 & 7 & $\{ 96,192,336,720\}$ & (34465, 11521, 11521) & 15min & Electricity \\
        
        \midrule
        Weather & 21 & $\{ 96,192,336,720\}$ & (36792, 5271, 10540) & 10min & Weather \\

        \midrule        
        ECL & 321 & $\{ 96,192,336,720\}$ & (18317, 2633, 5261) & Hourly & Electricity \\
        \hline
    \end{tabular}
    \caption{Detailed dataset descriptions.}
    \label{tab:Dataset}
\end{table*}

\begin{table*}[!t]
    \centering
    \begin{tabular}{c|c c|c c}
    \toprule
        Dataset & \multicolumn{2}{c}{ETTh1} & \multicolumn{2}{c}{ETTh2} \\
        \cmidrule(r){2-5}
        Metric & MSE & MAE & MSE & MAE \\
        \midrule
        96 & 0.371$\pm$0.002 & 0.397$\pm$0.001 & 0.290 $\pm$ 0.004 & 0.343 $\pm$ 0.003 \\
        \midrule
        192 & 0.434$\pm$0.006 & 0.429$\pm$0.002 & 0.370 $\pm$ 0.004 &  0.391 $\pm$ 0.002 \\
        \midrule
        336 & 0.467$\pm$0.002 & 0.446$\pm$0.001 & 0.420 $\pm$ 0.008 & 0.426 $\pm$ 0.004 \\
        \midrule
        720 &  0.468$\pm$0.012 & 0.466$\pm$0.005& 0.423 $\pm$ 0.006  & 0.435 $\pm$ 0.004 \\
        \toprule
        Dataset & \multicolumn{2}{c}{ETTm1} & \multicolumn{2}{c}{ETTm2} \\
        \cmidrule(r){2-5}
        Metric & MSE & MAE & MSE & MAE \\
        \midrule
        96 & 0.313$\pm$0.002 & 0.353 $\pm$ 0.003 & 0.172 $\pm$ 0.001 & 0.253 $\pm$ 0.001 \\
        \midrule
        192 & 0.359 $\pm$ 0.001 & 0.378 $\pm$ 0.001 & 0.236 $\pm$ 0.001 & 0.295 $\pm$ 0.001 \\
        \midrule
        336 & 0.389 $\pm$ 0.001 & 0.400 $\pm$ 0.001 & 0.292 $\pm$ 0.001 & 0.331 $\pm$ 0.001 \\
        \midrule
        720 & 0.459 $\pm$ 0.004 & 0.445 $\pm$ 0.002 & 0.394 $\pm$ 0.001 & 0.391 $\pm$ 0.001 \\
        \hline
    \end{tabular}
    \caption{Mean values and standard deviations of KFS.}
    \label{tab:Dataset}
\end{table*}

\subsubsection{Baseline Models}
We briefly describe the selected baselines:

TimeXer~\cite{wang2024timexer} TimeXer is a Transformer-based model that incorporates exogenous variables as supplementary inputs, achieving their integration with input variables through a cross-attention mechanism.

PatchTST~\cite{nie2023a} is a Transformer-based model utilizing patching and CI technique. It also enable effective pretraining and transfer learning across datasets.

iTransformer~\cite{liu2024itransformer} embeds each time series as variate tokens and is a fundamental backbone for time series forecasting.

TimesNet~\cite{wu2023timesnet}  is a CNN-based model with TimesBlock as a task-general backbone. It transforms 1D time series into 2D tensors to capture intraperiod and interperiod variations.

MICN~\cite{wang2023micn} is a CNN-based model combining local features and global correlations to capture the overall view of time series.

TimeMixer~\cite{wang2024timemixer} is an MLP-based model introducing multiscale mixing technology. It achieves a complex temporal pattern representation with seasonal trend decomposition.

DLinear~\cite{Zeng2022AreTE} is an MLP-based model with just one linear layer, which unexpectedly outperforms transformer-based models in long-term TSF.

FiLM~\cite{zhou2022film} is an MLP-based model. It conducts temporal pattern representation by legendre polynomials projections and fourier projection.

 Time-FFM~\cite{liu2024timeffm} is foundation model using LLM. It aligns modality representation and is empowered by language model.
\subsubsection{Metric Details}

Regarding metrics, we utilize the mean square error (MSE)
and mean absolute error (MAE) for long-term forecasting.
The calculations of these metrics are:
$$MSE=\frac{1}{T}\sum_{i=0}^T(\hat{y_i}-y_i)^2$$
$$MSE=\frac{1}{T}\sum_{i=0}^T|\hat{y_i}-y_i|$$

\subsubsection{Hyper-Parameter Selection and Implementation Details}

For the main experiments, we have the following hyperparameters. The dimension of embedding $d_{model}$. The hidden state of KAN $d_{ff}$. $d_{model}$ and $d_{ff}$ are set via hyperparameter searching among the range of $\{128, 256, 512, 1024\}$ for $d_{model}$ and $\{256, 512, 1024, 2048\}$ for $d_{ff}$. And we set $\alpha=0.3$ and $\delta=80\% $ in the ETTm1, ETTm2, Weather and ECL datasets. For ETTh1 and ETTh2, we search $\alpha$ in the range of 0 to 1 with step 0.1. And for $\delta$, the step is 0.05.  Detailed hyperparameters can be found at the Code \& Data Appendix.

\subsection{}section{Extra Experimental Results}
\subsubsection{}subsection{Robustness Evaluation}
To get more robust experimental results, we repeat each experiment three times with five seeds (2020-2024) in the ETTh1, ETTh2, ETTm1, ETTm2 datasets, demonstrating that KFS performance is stable. For easier comparison, the results are shown in the \textbf{main text} when the seed is set to 2021. All experiments are conducted using PyTorch on an NVIDIA 4090 24GB GPU and are repeated three times for consistency.

\subsection{Discussions on Limitations and Future Improvement}
Recently, several specific designs have been utilized to better capture complex sequential dynamics, such as RevIN, Frequency Representation, KAN method and Channel Independence.

\begin{figure*}
    \centering
    \includegraphics[width=1\linewidth]{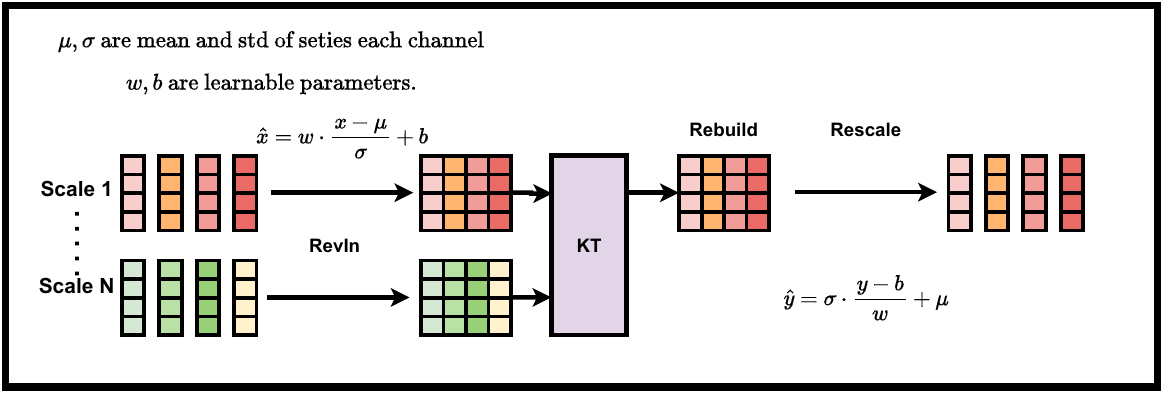}
    \caption{The Design for Sequential Modelling (RevIn)}
    \label{fig:Revin}
\end{figure*}

(1)RevIN: Real-world time series always exhibit non-stationary behavior, where the data distribution changes
over time. RevIn is a normalization-and-denormalization method for TSF to effectively constrain non-stationary information (mean and variance) in the input layer and restore it in the output layer. The work has managed to improve the delineation of temporal dependencies while minimizing the influence of noise. In KFS, we also adopt the method of RevIN. It is used for normalizing each scaled data, respectively (Figure~\ref{fig:Revin}. And the output is denormalized by the RevIN layer for original scale. However, it is difficult to adequately address the intricate distribution shifts between layers within deep networks, necessitating further refinement to resolve such shifts.

\begin{figure*}
    \centering
    \includegraphics[width=1\linewidth]{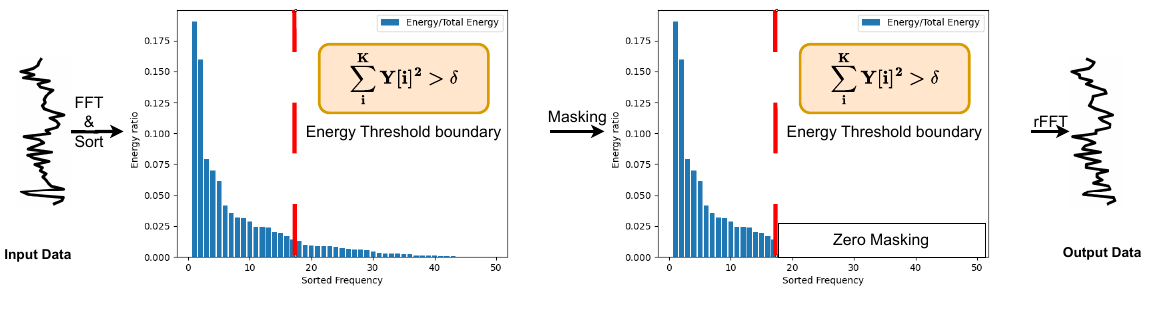}
    \caption{The Top-K Selection in our work.The sort operation is only used for search the index of frequency.}
    \label{fig:Energy}
\end{figure*}

(2)Frequency representation: TS data, characterized by their inherent complexity and dynamic nature, often contain information that is sparse and dispersed across the time domain. The frequency domain representation is proposed to promise a more compact and efficient representation of the inherent patterns. KFS minimizes the impact of noise via Frequency Band Selection showed in Figure~\ref{fig:Energy}. However,  some overlooked periods or trend changes may represent significant events, resulting in loss of information. 

(3)KAN method: KAN have emerged as a promising candidate to potentially replace MLPs, demonstrating exceptional representational capabilities. However, their substantial computational overhead and relatively homogeneous base functions have constrained their applicability to diverse domains, prompting the development of numerous variants. While these variants have achieved notable successes in their respective fields, there is no existing work that has yet designed efficient base functions specifically optimized for time series representation. Our KFS framework employs rational KAN as its feature representation backbone. Consequently, our future research will focus on developing domain-specific KAN variants grounded in the intrinsic mechanisms of time series data.

(4) Channel Independence (CI): the CI method sacrifices capacity in favor of more reliable predictions. PatchTST~\cite{nie2023a} achieves SOTA results using the CI approach. However, neglecting correlations between channels may lead to incomplete modeling. In KFS, we leverage the CI approach without integrating dependencies across different variables over time. However, in multivariate forecasting modeling, solely considering channel-wise information is incomplete as it neglects inter-channel dependencies. On the other hand, the complex trade-off between channel-specific patterns and cross-channel relationships complicates the integration mechanisms within models. Therefore, our future work will further explore channel correlation pattern discovery and effective fusion methods for diverse information sources.

We believe that more effective sequence modeling models will be proposed to adequately address issues such as distribution shift, multivariate sequence modeling, etc. As a result, KAN-based models have great potential waiting for further exploration in more areas of time series.


\end{document}